\documentclass[pdflatex,sn-mathphys-num]{sn-jnl}

\usepackage{graphicx}%
\usepackage{multirow}%
\usepackage{amsmath,amssymb,amsfonts}%
\usepackage{amsthm}%
\usepackage{mathrsfs}%
\usepackage[title]{appendix}%
\usepackage{xcolor}%
\usepackage{textcomp}%
\usepackage{manyfoot}%
\usepackage{booktabs}%
\usepackage{algorithm}%
\usepackage{algorithmicx}%
\usepackage{algpseudocode}%
\usepackage{listings}%
\usepackage{tikz}
\usepackage{tabularx}
\usepackage{booktabs, makecell, multirow, tabularx}
\newcommand*{\GridSize}{4}

\newcommand*{\ColorCells}[1]{
  \foreach \x/\y/\color in {#1} {
    \node [fill=\color, draw=none, thick, minimum size=1cm] 
      at (\x-.5,\GridSize+0.5-\y) {};
    }%
}%

\newcommand*{\Agent}[1]{
  \foreach \x/\y/\color in {#1} {
  \filldraw[color=\color!60, fill=\color!10, very thick](\x-.5,\GridSize+0.5-\y)circle (0.25);
   }%
}%

\theoremstyle{thmstyleone}%
\theoremstyle{thmstyletwo}%

\theoremstyle{thmstylethree}%

\begin{document}

\title[Article Title]{A Scalable Post-Processing Pipeline for Large-Scale Free-Space Multi-Agent Path Planning with PiBT}


\author*[1,2]{\fnm{Arjo} \sur{Chakravarty}}\email{arjoc@intrinsic.ai}

\author[1]{\fnm{Michael X.} \sur{Grey}}\email{mxgrey@intrinsic.ai}

\author[2]{\fnm{M. A. Viraj J. Muthugala} \sur{}}\email{viraj\_jagathpriya@sutd.edu.sg}

\author[2]{\fnm{Mohan Rajesh} \sur{Elara}}\email{rajeshelara@sutd.edu.sg}
\affil*[1]{\orgname{Intrinsic Innovation LLC}, \orgaddress{\street{3 Fusionopolis Wy, No. 13-21 Symbiosis}, \city{Singapore}, \postcode{138633}, \country{Singapore}}}

\affil[2]{\orgdiv{ROAR Lab}, \orgname{Singapore University of Technology and Design}, \orgaddress{\street{8 Somapah Road}, \city{Singapore}, \postcode{487370},  \country{Singapore}}}


\abstract{Free-space multi-agent path planning remains challenging at large scales. Most existing methods either offer optimality guarantees but do not scale beyond a few dozen agents, or rely on grid-world assumptions that do not generalize well to continuous space. In this work, we propose a hybrid, rule-based planning framework that combines Priority Inheritance with Backtracking (PiBT) with a novel safety-aware path smoothing method. Our approach extends PiBT to 8-connected grids and selectively applies string-pulling based smoothing while preserving collision safety through local interaction awareness and a fallback collision resolution step based on Safe Interval Path Planning (SIPP). This design allows us to reduce overall path lengths while maintaining real-time performance. We demonstrate that our method can scale to over 500 agents in large free-space environments, outperforming existing any-angle and optimal methods in terms of runtime, while producing near-optimal trajectories in sparse domains. Our results suggest this framework is a promising building block for scalable, real-time multi-agent navigation in robotics systems operating beyond grid constraints.}




\maketitle
\section{INTRODUCTION}

This paper explores the idea of a practical free-space multi-agent planner which scales to 100s of robots. We build on Priority inheritance BackTracking (PiBT) \cite{okumura2022priority}, which is currently one of the leading rules-based configuration generators for moving multiple agents. Thus far PiBT has largely been used in traditional grid-like Multi-agent Path Finding (MAPF) settings. Here we ask if it is possible to extend PiBT to free space planning. We combine path-smoothing, and Safe-Interval Path Planning (SIPP) with PiBT to introduce a new rules-based planner. While we do not provide any optimality guarantees, we show that the planner is able to find solutions significantly faster than Any-Angle Continuous-time Conflict Based Search (AA-CCBS) \cite{10801691}, which proposed the first any-angle planner. To do so, we break down this work into two pieces; first we simply extend PiBT from a 4-connected domain to an 8-connected domain. We empirically show that this reduces path lengths. Then we apply string-pulling to smooth the paths. This approach has similarities to navigation frameworks like Nav2 \cite{macenski2020marathon2}, which do not always need to use optimal any-angle planning. Our primary contribution is a novel path smoothing algorithm that provides safe free-space motion plans for a plan generated by PiBT.

\section{RELATED WORK\label{sec:lit_review}}

\subsection{Multi-Agent Path Finding (MAPF)}
Multi-Agent Path Finding (MAPF) is a vibrant domain. A lot of work has gone into solving MAPF optimally. Specifically, the community has targeted grid-like settings as these are good enough approximations of the real world warehouse settings and a good starting point for research. Currently there are several families of MAPF algorithms that can be broadly broken down into the following:
\begin{itemize}
    \item Conflict-Based Search (CBS) \cite{SHARON201540}: These algorithms rely on a two-level search. The low level search runs traditional path finding algorithms. When a conflict arises the high level search creates child nodes in a tree that split on the specific conflict. This approach has been shown to be very effective in the 10-100 robots setting \cite{li2021eecbs}. Additionally, its simplicity has lent itself to many extensions \cite{li2021eecbs}\cite{shaoul2024multi}. However, it often suffers from exploring too many conflicts and may have unpredictable runtimes \cite{GordonFS21}\cite{okumura2023lacam}.
    \item Compilation Based: Compilation based methods decompose the MAPF problem to equivalent NP-Hard problems such as MILP \cite{surynek2022problem} or SAT \cite{surynek2018sub}. This approach has limited scalability and is hard to extend \cite{surynek2022problem}.
    \item Rules Based: This category of solvers relies on bespoke hand-engineered heuristics. A recent breakthrough by Keisuke et al. \cite{okumura2022priority} has introduced PiBT as a viable rule based approach to multi-agent solving with practical scalability.
    \item Priority Based: These are the simplest solvers and scale well, however they often run into deadlocks \cite{okumura2022priority}.
    \item Learning Based: Learning based approaches such as PRIMAL \cite{sartoretti2019primal}, SCRIMP \cite{wang2023scrimp} and MAPF-GPT \cite{andreychuk2025mapf} are growing in popularity in part due to the recent advances in Deep Reinforcement Learning (RL) and improved GPU performance. Most Learning Based planners do need to be bootstrapped via imitation learning from other types of planners, and then they refine their strategies via RL. They are useful for decentralized, partially observable local behaviors \cite{andreychuk2025mapf}\cite{sartoretti2019primal}\cite{wang2023scrimp}.
\end{itemize}
Aside from these broad categories there are many varieties of planners that explore hybrid strategies. We choose to work with PiBT as it can be used as a component by other planners and provides unprecedented scalability.

Grid-world representations have been a central focus of MAPF research. Several standard benchmarks for these exist \cite{chan2024league}\cite{stern2019mapf}. These provide an excellent but simple base to research new search schemes. However, many real world robotics scenarios would benefit from not being constrained to the restrictive 4-connected grid approach. Adreychuk et al. \cite{10801691} proposed AA-CCBS which builds on CCBS and combines itself with SIPP to achieve optimal any-angle planning. A similar approach was attempted by Open-RMF's MAPF library as well \cite{githubGitHubOpenrmfmapf}. While these provide optimality guarantees, many applications prefer a solution to be available (even if suboptimal) than having to wait for a solver to arrive at the optimal solution.

PIBT was proposed by Keisuke Et Al. as a ``Configuration Generator" for MAPF solvers. It provides a certain simple set of rules that gives feasible configurations quickly. PiBT combines the notion of priority based planning together with deadlock resolution methods. Specifically, PiBT uses priority inheritance to prevent priority inversion and backtracking to resolve deadlocks. It has been proven to be complete for the Multi-Agent Pickup and Delivery (MAPD) problem on bi-connected graphs \cite{okumura2022priority}. Its speed makes it practical for many modern MAPF planners to use it as a low-level planner \cite{MAPFLNS_2022}\cite{okumura2023lacam}.

\subsection{Path Smoothing and Single-Agent Navigation}

When dealing with free-space navigation, theta* is commonly used for any-angle path finding \cite{nash2007theta}. However, theta* can get expensive as it needs to perform line of sight checks \cite{nash2010lazy}. Several approaches to solve this have been proposed, including lazy-theta* \cite{nash2010lazy}. For our purpose, it is more important to be able to scale to more robots than to provide perfect optimality. This is important for large-scale applications as it is an established fact that optimal MAPF is NP-hard. Suboptimal MAPF is believed to be solvable in poly-time. In fact, PiBT is known to have $O(|A|(\Delta (G) + log A))$ time complexity \cite{okumura2022priority} per timestep. In conventional navigation, like Nav2, it is a common practice to use path planners that do not guarantee any-angle optimality and then smooth the paths \cite{macenski2020marathon2} post-hoc. Some common path smoothers include Savitzky-Golay, string pulling and constraint path smoothing \cite{macenski2020marathon2}. The challenge with path smoothing is its inherent limitations, as it frequently introduces unintended side effects. Particularly in the context of MAPF, smoothing often causes path deviations which may lead to collisions or deadlocks. We propose a procedure that significantly reduces the number of intersections in comparison to traditional path smoothing techniques. 

\subsection{Eight Connected Grid - A Good Enough Approximation}

There are many ways to represent a free space. In the domain of single agent planning, it is common to use a grid. When planning for one agent, visibility graphs or theta* can accelerate optimal path finding. However, the improvement that theta* has over classic 8-connected grids is roughly 8\% \cite{Nash2012}. 

\section{Proposed Algorithm}

We propose a path-smoother that guarantees collision-free paths while reusing the results of eight connected PiBT. 

\subsection{Eight connected PiBT}

\begin{figure}[hp]
\centering
\includegraphics[width=0.8\linewidth]{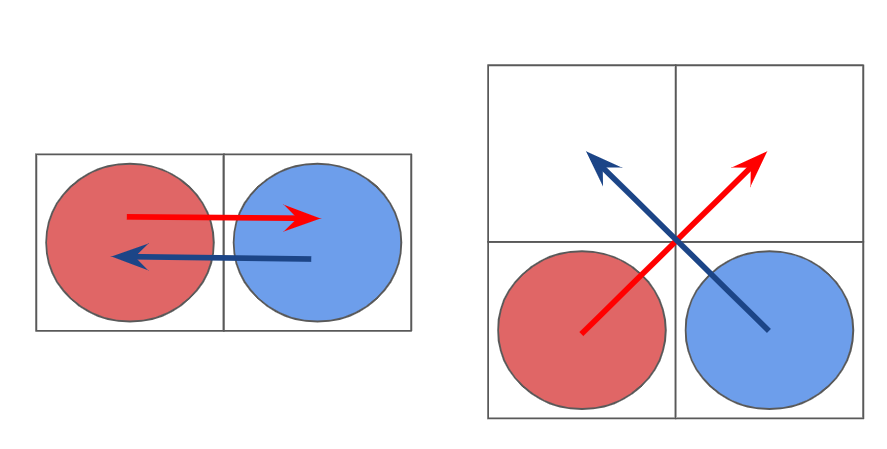}
    \caption{On the left we have the traditional swap conflict that is handled by classic PiBT. On the right we have additional swap conflicts that needs to be handled in 8-connect PiBT.}
    \label{fig:swap_conf}
\end{figure}

There are two possible ways to extend PiBT to the 8-connected case: The first is to simply extend the ``get neighbors'' function and update the collision check to take into account the diagonal cases. This is shown by Figure \ref{fig:swap_conf}. The second more complicated way would be to extend priority inheritance over multiple cells. For this work we only consider the first approach. 

Despite a speed penalty for the higher branching factor, we find 8-connect PiBT is able to scale to 100s of robots. Additionally, we find paths proposed by 8-connected PiBT to be shorter than their 4-connected counter parts. One of the pit-falls of 8 connected PiBT however is that the travel time in the diagonal is not the same as the travel time in cardinal directions. For the purpose of this paper, we assume the robots can operate at a range of speeds, with the maximum speed being more than the time it takes to cover a diagonal of a single cell. It can be shown that any line that is contained in a square has to be shorter or equal to the length of its diagonal thus, we can use this as the max robot speed. Using 8-connect PiBT is important as when we perform string pulling, the newly formed line that shortens the path can deviate significantly from the original path. With the 4-connect variant, this can be problematic as we may aggressively pull strings leading to more collisions (see Section \ref{sec:benchmark}). 

\subsection{Safety-aware string pulling for Path Smoothing}
\begin{figure}[hp]
\centering
\includegraphics[width=0.8\linewidth]{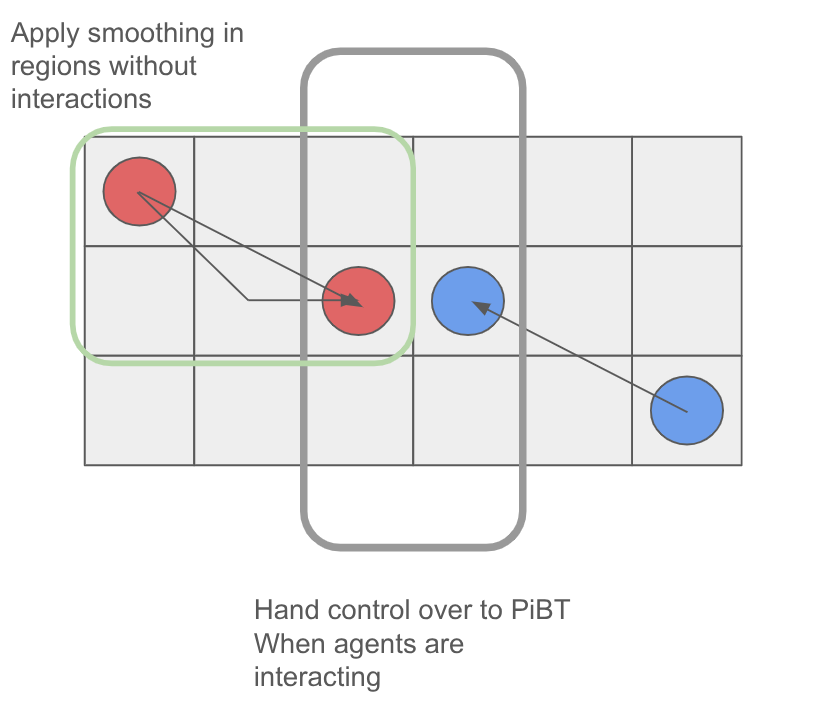}
    \caption{Our key insight is that since interactions in PiBT are local path smoothing should happen only when agents are far from each other.}
    \label{fig:key_insight}
\end{figure}
\begin{figure}
\centering
\begin{tikzpicture}[scale=1]

    \begin{scope}[thick,local bounding box=name]
        \ColorCells{1/1/red!10, 1/2/red!10, 2/1/red!10, 2/2/red!10, 4/4/blue!10, 3/3/blue!10, 3/4/blue!10, 4/3/blue!10}
        \Agent{1/1/red,4/4/blue}
        \draw (0, 0) grid (\GridSize, \GridSize);
    \end{scope}

\end{tikzpicture}
\caption{Areas of critical interaction: In this case the blue agent can move into any of the blue colored squares. Other agents should avoid pulling into these regions.}
\label{fig:crit_int}
\end{figure}
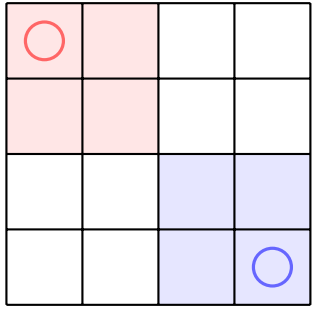

The key insight our algorithm uses is that PiBT negotiations are local, hence we hand control to PiBT when agents are \textit{Directly Interacting} and hand control to our smoother when agents are not directly interacting as illustrated in Figure \ref{fig:key_insight}.   

String pulling is a classic technique for shortening paths produced by A* \cite{cui2011direction}. However, when dealing with free space planning and the output of PiBT, it is important to take into account how agents interact. In this section we directly borrow terminology from the original PiBT paper \cite{okumura2022priority}. The interaction of agents is governed by the notion of \textit{interacting agents}. \textit{Interacting agents} are defined by the fact that PiBT is a local-first configuration generator. If two agents are located within two hops of each other they are considered to be \textit{directly interacting}. A group of \textit{Interacting agents} are defined transitively over chains of directly interacting agents. It follows that every \textit{interacting agent} must be \textit{directly interacting} with some agent. Therefore, a simple space-time search should reveal neighbors of an agent. If there are interactions, we do not attempt to smooth them. This alone does not guarantee safety of the newly smoothed path. We must ensure that our pulled-string does not cross into regions of \textit{critical interactions}. We define an area of \textit{critical interaction} as an area that is within 1 hop of another agent as the agent may enter this region in the next time step and was not expecting an obstacle (See Figure \ref{fig:crit_int}). 

We refer our reader to Algorithm \ref{alg:alg-pibt}.  If an agent is an \textit{interacting agent} we make sure the agent reaches that critical interaction at the allotted time and do not perform any smoothing while agents are in critical interactions (See line \ref{line:break} of Algorithm \ref{alg:alg-pibt}). Our line of sight check during string pulling takes into account regions of \textit{Critical Interaction} by treating such regions as a time bound obstacle (line \ref{line:los_check}).

Even after these restrictions, we do not have a guaranteed collision free path. It is possible that 2 newly pulled strings might intersect at the same time and place. To avoid this, we rely on a complementary solver. In this case, we implement SIPP with a priority based scheme to prevent the two agents from colliding. In our benchmarks we found that the number of intersections were low, and all of them were possible to resolve using SIPP. However, a more complete solution would be an interesting direction for future work.

\begin{algorithm}[hp]
\caption{PIBT Path Smoothing}
\label{alg:alg-pibt}relatively rare
\textbf{Parameters}: PiBT Path Smoothing \\ \noindent
\textbf{Output}: Collision free smoothed path $a^{1:N}$
\begin{algorithmic}[1] 
\Procedure{String-Pull}{$T$, $O$, $C^{1:N}(t)$}
    \State $T$: the trajectory consisting of tuples $(t_i, x_i, y_i)$
    \State $O$: the list of obstacles
    \State $C$: a list of regions of critical interaction
    \State $T_{new} = \{T(0)\}$
    \For{$t_i < t_{MAX}$}
        \For{$t_j \in [t_{i+1}..t_{MAX})$}
            \State $t_{next} \gets t_j$ 
            \If{$\neg$line\_of\_sight$(T[t_i],T[t_j],O,C)$ } \label{line:los_check}
               \State break
            \EndIf
        \EndFor
        \State $T_{i} \gets$ \Call{LinearInterpolation}{$T[t_i]$, $T[t_{next}]$}
        \State \Call{Append}{$T_{new}$, $T_{i}$}
        \State $t_i \gets t_{next}$ 
    \EndFor
    \State \textbf{return} $T_{new}$
\EndProcedure
\Procedure{PIBT-Smoother}{$s^{1:N}(t)$, $O$}
    \State $s^{1:N}(t)$: Agent trajectories from PiBT solver
    \State $O$: List of obstacles
    \State $ C^{1:N}(t) \gets s^{1:N}(0)$
    \For{time $t \in 0..t_{MAX}$} \label{line:priorities}
        \For{agent $n \in 1...N$}
        \If{\Call{CriticalInteraction}{$s^n(t)$}}
            \State \Call{Append}{$C^n(t)$, $s^n(t)$}
        \EndIf
        \EndFor
    \EndFor
    \State $ C^{1:N}(t_{MAX}) \gets s^{1:N}(t_{MAX})$

    \For{agent $n \in 1...N$} \label{line:break}
        \For{$s_0, s_1 \in \Call{ConsecutivePairs}{C^n}$}
        \State $T^n \gets$ \Call{TrajectorySegment}{$s_0$, $s_1$}
        \State $T^n \gets$ \Call{StringPull}{$T^n$, $O$, $C^{1:N}$}
        \EndFor
    \EndFor

    \For{agent $n_1 \in 1...N$}
        \For{agent $n_2 \in n_1+1...N$}
        \If{\Call{Intersection}{$T^{n_1}$, $T^{n_2}$}}
            \State \Call{ApplySIPP}{$T^{n_1}$, $T^{n_2}$}
        \EndIf
        \EndFor
    \EndFor

    \State \textbf{return} 
\EndProcedure
\end{algorithmic}
\end{algorithm}
\section{Benchmarks}
\label{sec:benchmark}

\begin{figure}[hp]
    \centering
    \includegraphics[width=0.8\linewidth]{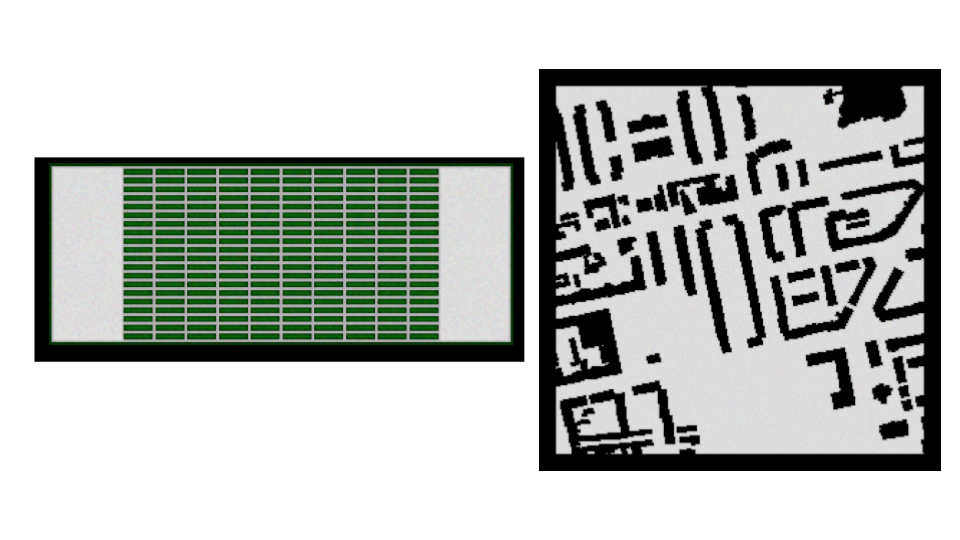}
    \caption{The two  maps we refer to most frequently in our paper. We ran experiments on all the other maps in the benchmarks, but frequently refer to these. On the left is Warehouse-20-4-10-2-1, on the right is Berlin\_256\_256. The names we use are derived from the original benchmark \cite{stern2019mapf}}
    \label{fig:maps}
\end{figure}

We have made available a Python implementation of our algorithm \footnote{\url{https://github.com/arjo129/pypibt}}. We benchmark this against CCBS+SIPP as implemented in Open-RMF's MAPF library \cite{githubGitHubOpenrmfmapf}. Our benchmarks use the MAPF benchmarks by Stern et al. \cite{stern2019mapf}. In our results table we reuse the same scenario names as those in \cite{stern2019mapf}. We benchmark the post-smoothing behavior against a pre-smoothing behavior and show our improvements. We also show the cost reduction as compared to CCBS+SIPP. All benchmarks were run on a Thinkpad P50 with a 16 core i7-11850H processor and 64GB RAM. We provide the raw benchmark results for 800 scenarios in the MAPF benchmark in a supplementary file. For the purpose of brevity in our discussion we will highlight the two maps in which we see the best performance (Berlin\_256\_256) and worst performance (Warehouse-20-4-10-2-1) for all our tables. These maps have been reproduced for the reader in Figure \ref{fig:maps}. They represent two extremes, the Warehouse-20-4-10-2-1 instance is extremely grid-like where the shortest paths are likely to follow the cardinal directions, whereas the Berlin\_256\_256 instance has lots of freespace enabling a large amount of optimization to take place. 

Table \ref{tab:soln_times} compares the solution times for a set of maps from 
the MAPF benchmark. As we can see CCBS+SIPP can scale to 4 robots easily but struggles when we reach the 10 robots range. On the other hand, we find that the PiBT+Path Smoothing approach is able to scale to the 100 robots range.

\begin{table}[hp]
    \centering
    \caption{Solution Times compared to CCBS + SIPP}
    \begin{tabular}{@{}cccc@{}} \toprule
         Scenario Name&   \makecell{Number of\\Agents}&  \makecell{Solve Time\\ CCBS+SIPP}& \makecell{Solve Time \\ PiBT+PS} \\ \midrule
         \multirow{4}{*}{\shortstack{warehouse-20-40-10-2-1\\random-3}}&  2&  0.249s& 0.165s\\  
         &  4&  0.595s& 0.547s\\  
         &  10&  2.15s& 1.55s\\ 
         &  100& Unsolved& 13.5s\\ \hline
         \multirow{3}{*}{\shortstack{Berlin\_1\_256\\random-20}}& 2& 0.28s & 0.29s \\ 
         & 4& Unsolved& 0.67s \\ 
         & 100& Unsolved& 27.8s \\ \botrule
    \end{tabular}
    
    \label{tab:soln_times}
\end{table}
Comparing AA-CCBS+SIPP with our path smoother presents inherent discrepancies, primarily because AA-CCBS+SIPP includes optimality guarantees, which our approach does not offer. Hence we compare the arclength of the generated paths in Table \ref{tab:arclength_comp}. This is important as while PiBT guarantees reachability on bi-connected graphs, there are no optimality guarantees provided by it. We find that despite this fact, PiBT+Path Smoothing on average produces paths that are 0.8\% worse than the CCBS+SIPP optimal approach across the entire mapf benchmark. There are several limitations to this statement. First, we only know the optimal freespace paths for few agents as often CCBS+SIPP could not scale beyond 4 robots. Thus, we feel Table \ref{tab:arclength_comp} is much more representative of the expected performance hit.

\begin{table}[hp]
    \centering
    \caption{Arclength compared to CCBS+SIPP}
    \begin{tabular}{@{}cccc@{}} \toprule
         Map Name&  Number of Agents &  Arclength CCBS+SIPP& Arclength PiBT+PS\\ \midrule 
         \multirow{4}{*}{\shortstack{warehouse-20-40-10-2-1\\random-3}}&  2&  256.4& 256.4\\  
         &  4&  730.4& 735.1\\  
         &  10&  1945.0& 1960.6\\  
         &  15&  2591.9& 2625.3\\ \hline
         \multirow{3}{*}{\shortstack{Berlin\_1\_256\\random-20}}& 2& 175& 178 \\  
         & 4& Unsolved& 440 \\  
         & 100& Unsolved& 16589.3 \\ \botrule
    \end{tabular}
    \label{tab:arclength_comp}
\end{table}

Table \ref{tab:solntime_pibt} shows the impact of our path smoothing in comparison to vanilla PiBT. The gap between vanilla PiBT and our path smoothing increases with the number of agents. This is shown even more clearly when dealing with large numbers of agents in Figure \ref{fig:path-smoother-perf}. Essentially for small number of agents the overhead of computing path smoothing is low, however as the numbers increase the overhead begins to consume a significant chunk of time.
\begin{table}[hp]
    \centering
    \caption{Solution Times With and Without Path Smoothing}
    \begin{tabular}{@{}cccc@{}} \toprule
         Map Name&  \makecell{Number of\\Agents}&  \makecell{8-connect\\PiBT}& \makecell{8-connect PiBT\\Path Smoothing} \\ \midrule
        \multirow{6}{*}{\shortstack{warehouse-20-40-10-2-1\\random-3}}&  2&  0.15s& 0.17s\\  
         &  4&  0.50s& 0.55s\\
         &  10&  1.4s& 1.5s\\ 
         &  15&  1.9s& 2.1s\\ 
         &  20&  2.2s& 2.5s\\ 
         &  100&  11.2s& 13.5s\\ \hline
         \multirow{6}{*}{\shortstack{Berlin\_1\_256\\random-20}}&  2& 0.28s& 0.30s \\ 
         &  4&  0.63s& 0.67s\\ 
         &  10&  2.5s& 2.6s\\  
         &  15&  3.9s& 4.1s\\ 
         &  20&  5.2s& 5.4s\\ 
         &  100&  25.7s& 27.8s\\ \botrule
    \end{tabular}
    \label{tab:solntime_pibt}
\end{table}
Table \ref{tab:colred} is an ablation study on taking away certain components and seeing how frequently we need to fallback to using some local collision resolution strategy. The first configuration we compare using 4-connect PiBT vs 8-connect PiBT for number of collisions. Additionally, we compare the ``Safety-Aware" version to a vanilla string pulling which ignores the \textit{Critical Regions}. We find that the \textit{Critical-Region} aware versions produce an order of magnitude less intersections, allowing us to minimize the number of times we need to call our fallback planner.
\begin{table}[hp]
    \centering
     \caption{Intersections introduced by path smoothing}
    \begin{tabular}{@{}ccccc@{}} \toprule
         Map Name&  \makecell{Number of\\Agents}&  \makecell{4-connect\\PiBT+PS}& \makecell{8-connect\\PiBT+PS} & \makecell{Ignoring\\critical regions} \\ \midrule 
         Berlin\_1\_256&  50&  3& 0& 13\\ 
         Berlin\_1\_256&  100&  6& 2& 63\\ 
         Berlin\_1\_256&  200&  12& 11& 170\\ 
         Berlin\_1\_256&  500&  25& 21& 980\\ 
         \botrule
    \end{tabular}
   
    \label{tab:colred}
\end{table}
Finally, we also benchmark the effectiveness of the path smoothing by showing the total path length reduction. Table \ref{tab:arclength} shows the path length reduction in the best case scenario. For 4 robots we can reduce the path length by as much as 27.8\%. The general trend is the greater the density of robots the less effective our path smoothing approach.

\begin{table}[hp]
    \centering
    \caption{Path length reduction vs Number of Agents}
    \begin{tabular}{@{}ccc@{}} \toprule
         Map Name&  \makecell{Number of\\Agents}&  \makecell{Path Length Reduction} \\ \midrule
         Berlin\_1\_256&  4&  27.8\%  \\ 
         Berlin\_1\_256&  10& 15.0\%  \\ 
         Berlin\_1\_256&  15& 14.3\%  \\ 
         Berlin\_1\_256&  20& 12.9\%  \\ 
         Berlin\_1\_256&  100& 10.0\%  \\ 
         \botrule
    \end{tabular}
    
    \label{tab:arclength}
\end{table}

\section{Discussion}
\begin{figure}[hp!]
    \centering
    \includegraphics[width=0.6\linewidth]{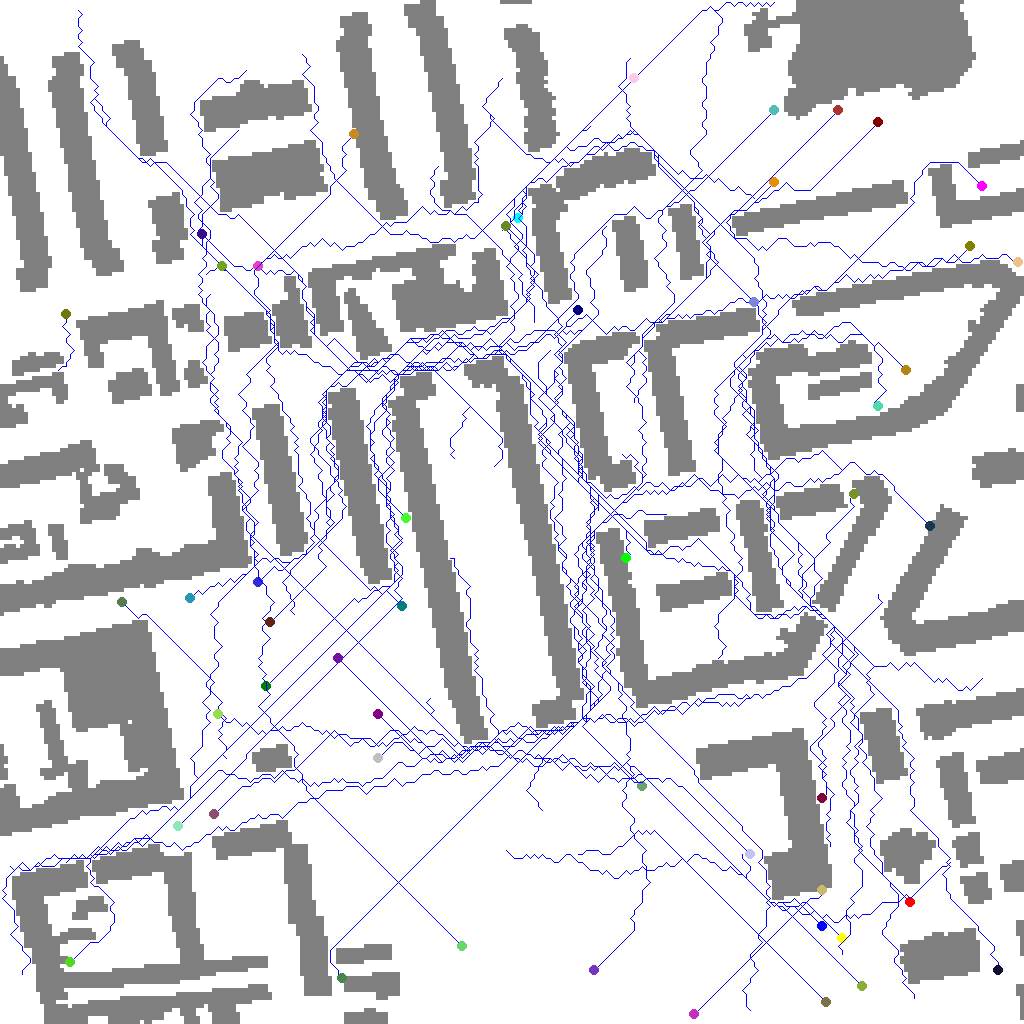}
    \caption{8-connect PiBT without path smoothing}
    \label{fig:no-pull}
\end{figure}

\begin{figure}[hp!]
    \centering
    \includegraphics[width=0.6\linewidth]{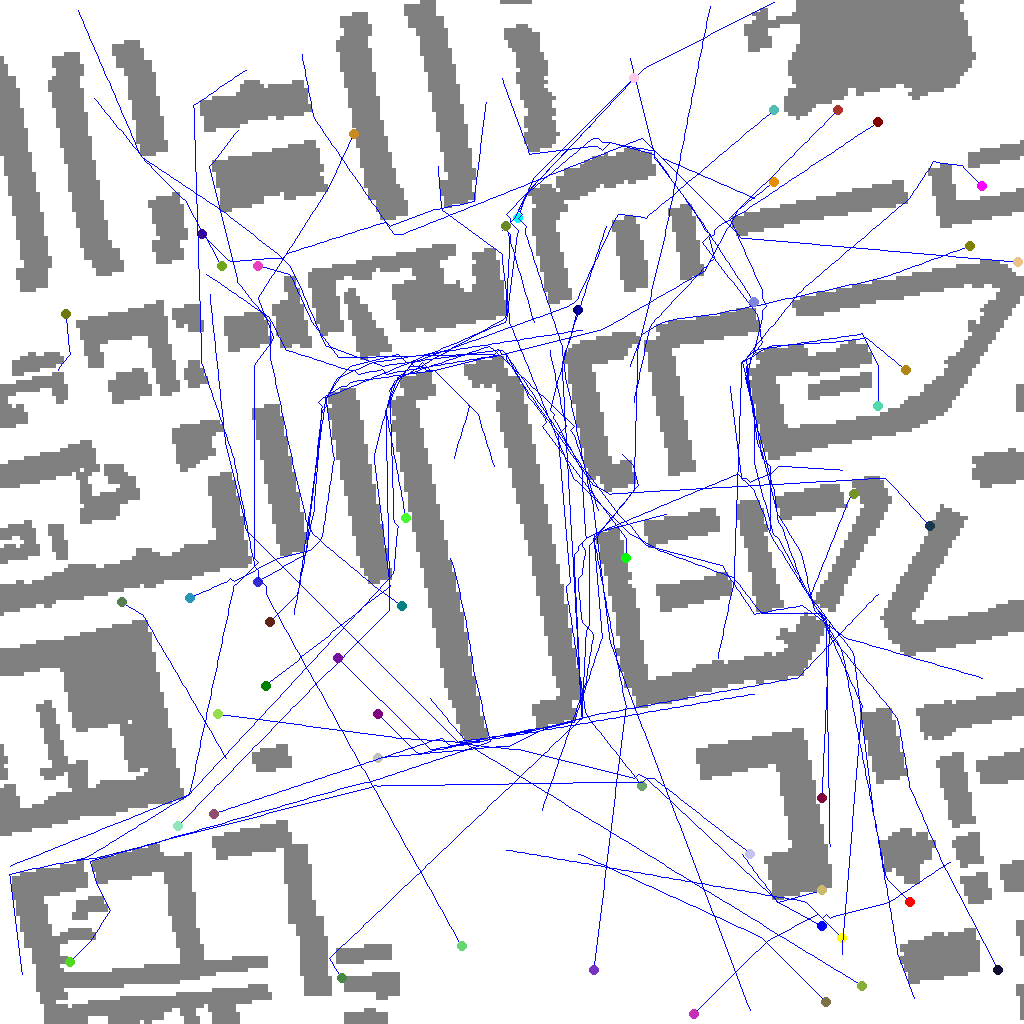}
    \caption{8-connect PiBT with string path smoothing. Notice the improved paths as compared to Figure \ref{fig:no-pull}.
    }
    \label{fig:yes-pull}
\end{figure}

\begin{figure}[!htb]
    \centering
    \includegraphics[width=0.6\linewidth]{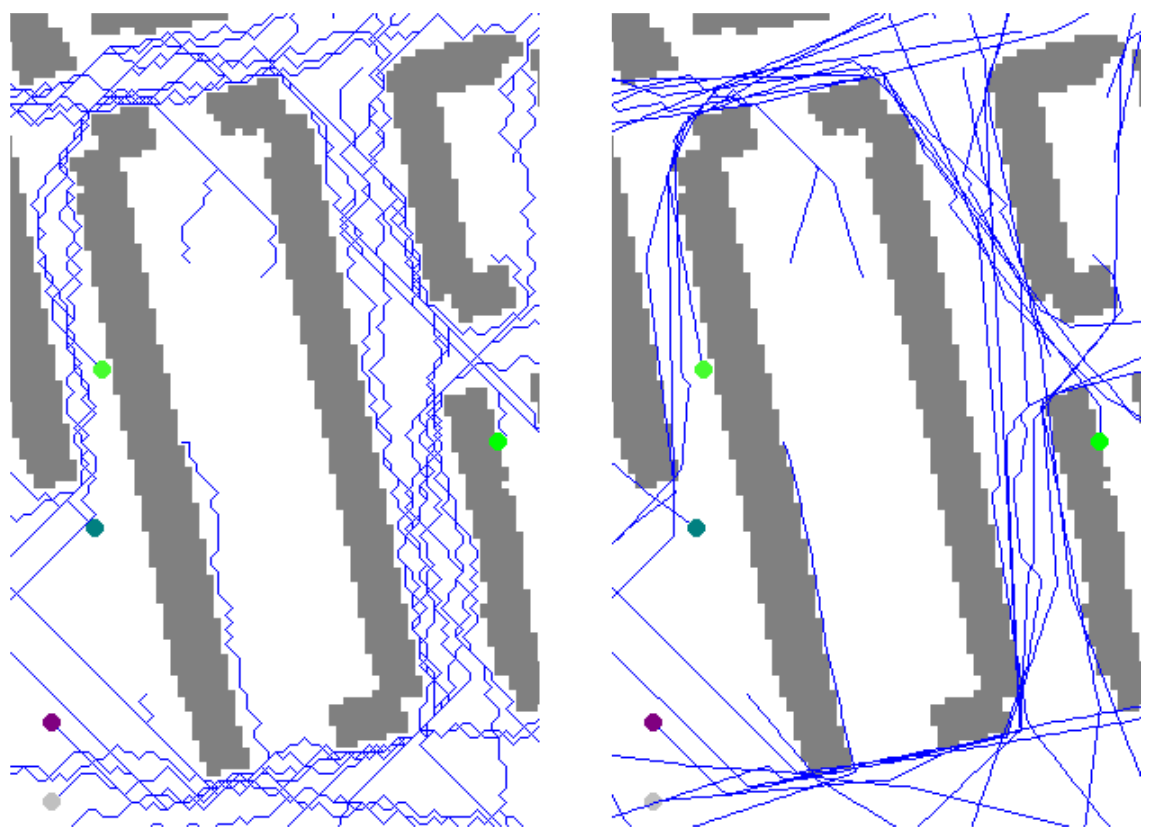}
    \caption{A zoomed view comparing the smoothing action in \ref{fig:no-pull} and \ref{fig:yes-pull}. The left side shows the unsmoothed version and the right shows the smoothed version.
    }
    \label{fig:side_by_side}
\end{figure}

Figure \ref{fig:no-pull} and Figure \ref{fig:yes-pull}, show the before and after effects of our path smoothing scheme. As can be seen in Figure \ref{fig:yes-pull}, the trajectories are less jagged than the trajectories in Figure \ref{fig:no-pull}. A close up zoom of the congested pathway is shown in \ref{fig:side_by_side} This is important when planning in physical space as frequent direction changes are not good for robots.

Compared to AA-CCBS+SIPP, we show that our PiBT based approach has the ability to scale more effectively. As seen in Table \ref{tab:soln_times} our solver can find solutions for 10s-100s of agents in a practical timeframe, whereas AA-CCBS+SIPP can achieve freespace planning for 2-15 agents. It is important to note that AA-CBS+SIPP offers optimality guarantees which we do not. However as the optimal variant of MAPF is known to be NP-hard \cite{conf/aaai/Surynek10}, in many real-world scenarios we would prefer methods with a bounded run-time.

The path smoothing approach is particularly effective when there are large empty areas with geometries that may not fit well into a grid world. This explains the extreme effectiveness in the Berlin map, where path smoothing was able to reduce path length by up to 27.8\% of the original path length proposed by PiBT.  On the flip side, the path smoothing has almost no effect in worlds with narrow grid patterns like the warehouse world where the improvements via path smoothing are as low as 0\%.

\begin{figure}[!htb]
    \centering
    \includegraphics[width=0.6\linewidth]{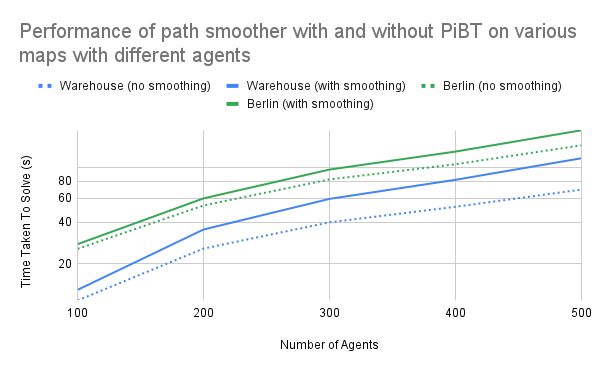}
    \caption{Performance of Smoother vs without Smoother on select maps in the 100s of robots range.}
    \label{fig:path-smoother-perf}
\end{figure}

The path smoothing approach is also not appropriate for reducing makespan as the makespan of the trajectories is preserved whenever there is any \textit{critical interaction}. A consequence of this is shown in Table \ref{tab:arclength} where the more the agents, the less effective the smoothing is at reducing path length. In very dense set-ups this may mean zero improvements over the original trajectories. Fortunately, the path smoothing procedure is light-weight. As shown in Figure \ref{fig:path-smoother-perf}, the smoother only consumes a small fraction of time that it requires to solve a mapf instance. It is also possible to parallelize the smoothing process per-agent, however that is not the focus of this work.

Perhaps the most significant drawback in our algorithm is the need to check for intersections and resolve them via traditional MAPF techniques at the end of our smoothing if new trajectory intersections arise. Our experiments show that such intersections do arise when handling hundreds of robots, however the number of new intersections stays well below 20 in most cases. Such intersections usually arise for a limited number of robots, so their resolution is relatively easy to compute and the solutions for each intersection can be computed independently from each other.

Based on our findings in Table \ref{tab:colred}, we see that considering areas of \textit{Critical Interaction} is incredibly important for avoiding the need to fallback to local collision avoidance techniques. 

Apart from the specific cases highlighted in our discussion, we observe an average pathlength reduction of 7.3\% across all tested scenarios and had an average overhead of 19\% for larger instances ($\geq$100 robots). The overhead was much less in smaller instances where it could be as low as 2\%.

\section{Conclusion and Future work}
We have shown a method for path smoothing over PiBT. Our key insight was that we could use the local properties of PiBT to reduce the number of collisions introduced when smoothing the path. By partitioning the path into segments where PiBT has full control due to high robot density and segments which are safe to smooth, we have shown that we can reduce path length in maps with large amounts of free space while introducing a minimal number of new trajectory intersections. This approach allows us to approximate free space traversal of robots at a much larger scale than conventional AA-CCBS+SIPP.

Several factors that may be relevant to real-world applications are not modeled in our method. First, we do not look at Kinodynamics of the robot. This can be problematic if the planner introduces a particularly sharp turn or for instance the robot is unable to come to an immediate stop. Second, our approach assumes homogeneity of robots. In reality not all robots will behave in the same way. Deployments may mix different
types of robots. This type of heterogeneity is not supported in PiBT and could be a good future direction for research.

We also cannot reduce the makespan of a smoothed path as we rely on the conflicts happening at fixed times dictated by the original plan. A good future direction would be to try to derive a semantic plan instead of a time based plan and then apply path-smoothing in a Post-Hoc fashion. This may enable us to reduce the total make-span.

Additionally, the escape hatch we use to avoid collisions at newly introduced trajectory intersections is less than ideal. Future work can focus on finding a way to ensure collisions do not take place to begin with. Alternatively instead of using SIPP\cite{phillips2011sipp}, other methods like Optimal Reciprocal Collision Avoidance (ORCA)\cite{alonso2013optimal}, Control Barrier Functions \cite{gao2023online} (or for that matter even PiBT) can be used to resolve such collisions.

Our work also makes no claims about optimality or completeness of our method. This is an important area for future research. Perhaps we can work such a path smoother into an optimizer like LaCAM* \cite{okumura2023lacam} in future to provide optimality guarantees.

\bmhead{Acknowledgements}
This research is supported by the National Robotics Programme under its National Robotics Programme (NRP) BAU, Ermine III: Deployable Reconfigurable Robots, Award No. M22NBK0054 and also supported by A*STAR under its RIE2025 IAF-PP programme, Modular Reconfigurable Mobile Robots (MR)2, Grant No. M24N2a0039.

\section*{Declarations}

\begin{itemize}
\item Funding: 
This research is supported by the National Robotics Programme under its National Robotics Programme (NRP) BAU, Ermine III: Deployable Reconfigurable Robots, Award No. M22NBK0054 and also supported by A*STAR under its RIE2025 IAF-PP programme, Modular Reconfigurable Mobile Robots (MR)2, Grant No. M24N2a0039.
\item Conflict of interest/Competing interests: The authors declare no conflicts of interest.
\item Ethics approval and consent to participate: Not applicable
\item Consent for publication: Not applicable
\item Data availability: Benchmark results are available here: \url{https://github.com/arjo129/pypibt/tree/main/docs/results}, the mapf scenario and files are available here: \url{https://movingai.com/benchmarks/mapf.html} 
\item Materials availability: See code availability
\item Code availability: \url{https://github.com/arjo129/pypibt}
\item Author contribution: \textbf{Arjo Chakravarty:} Conceptualization, Software, Formal Analysis, Validation, Writing – Original Draft, \textbf{M. X. Grey:} Conceptualization, Validation, Supervision, Writing – Review \& Editing, \textbf{Viraj:} Validation, Supervision, Writing – Review \& Editing, \textbf{Mohan Rajesh Elara:} Validation, Supervision, Funding Acquisition
\end{itemize}

\bibliography{sn-bibliography}

\end{document}